\documentclass[review]{elsarticle}
\graphicspath{ {./figures/} }
\usepackage{hyperref}
\usepackage{float}
\usepackage{verbatim} 
\usepackage{apalike}
\restylefloat{figure}
\usepackage{rotating} 

\graphicspath{ {./figures/} }
\usepackage[table]{xcolor}
\usepackage{verbatim} 
\usepackage{apalike}
\usepackage{caption}
\usepackage{subcaption}
\usepackage[margin=1in]{geometry} 
\usepackage[]{graphicx}
\usepackage[font=small,skip=0pt]{caption}
\usepackage{enumitem}
\usepackage{booktabs}
\usepackage{longtable}
\usepackage{tabularx}
\usepackage{cancel}
\usepackage{multirow}
\usepackage{supertabular}
\usepackage{algorithmic}
\usepackage{algorithm}
\usepackage{amsmath}
\usepackage{rotating}
\usepackage[normalem]{ulem}
\usepackage[utf8]{inputenc}
\usepackage[T1]{fontenc}
\usepackage{lineno}
\usepackage{array}
\usepackage{makecell}
\usepackage{pdflscape}
\usepackage{afterpage}
\usepackage{capt-of}
\usepackage{lipsum}
\usepackage{changepage}
\usepackage{xcolor}
\usepackage{rotating} 
\usepackage{graphicx}
\newcommand{\blue}[1]{\textcolor{blue}{#1}}
\usepackage{amssymb}
\usepackage{natbib}
\bibpunct{(}{)}{,}{a}{}{;}
\renewcommand{\citep}[1]{(\citealp{#1})}

\journal{JAG International Journal of Applied Earth Observation and Geoinformation}

\biboptions{authoryear}

\begin{document}

\begin{frontmatter}
\begin{titlepage}
\begin{center}
\vspace*{1cm}

\textbf{ \large Individual mapping of large polymorphic shrubs in high mountains using satellite images and deep learning}

\vspace{1.5cm}

Rohaifa Khaldi$^{a, b, e}$ (rohaifa@go.ugr.es), Siham Tabik$^b$ (siham@ugr.es), Sergio Puertas-Ruiz$^f$ (s.p.r@csic.es), Julio Peñas de Giles$^c$ (jgiles@ugr.es), José Antonio Hódar Correa$^c$ (jhodar@ugr.es), Regino Zamora$^{d, e}$ (rzamora@ugr.es), Domingo Alcaraz Segura$^{c, e}$ (dalcaraz@ugr.es) \\

\hspace{10pt}

\begin{flushleft}
\small  
$^a$ LifeWatch-ERIC, ICT Core, 41071 Seville, Spain. \\
$^b$ Dept. of Computer Science and Artificial Intelligence, Andalusian Research Institute in Data Science and Computational Intelligence, DaSCI, University of Granada, 18071 Granada, Spain. \\
$^c$ Dept. of Botany, Faculty of Science, University of Granada, 18071 Granada, Spain. \\
$^d$ Dept. of Ecology, Faculty of Science, University of Granada, 18071 Granada, Spain. \\
$^e$ Interuniversity Institute of Earth System Research in Andalusia, Andalusian Center for the Environment (IISTA-CEAMA), Granada, 18071, Spain \\
$^f$ Pyrenean Institute of Ecology, Spanish National Research Council (IPE-CSIC), 50059 Zaragoza, Spain.

\end{flushleft}        
\end{center}
\end{titlepage}

\title{Individual mapping of large polymorphic shrubs in high mountains using satellite images and deep learning}

\author[label1,label2,label5]{Rohaifa Khaldi \corref{cor1}}
\ead{rohaifa@ugr.es}
\author[label2]{Siham Tabik \corref{cor1}}
\ead{siham@ugr.es}
\author[label6]{Sergio Puertas-Ruiz}
\ead{s.p.r@csic.es}
\author[label3]{Julio Peñas de Giles}
\ead{jgiles@ugr.es}
\author[label3]{José Antonio Hódar Correa}
\ead{jhodar@ugr.es}
\author[label4,label5]{Regino Zamora \corref{cor1}}
\ead{rzamora@ugr.es}
\author[label3,label5]{Domingo Alcaraz Segura \corref{cor1}}
\ead{dalcaraz@ugr.es}

\cortext[cor1]{Corresponding author.}
\address[label1]{LifeWatch-ERIC, ICT Core, 41071 Seville, Spain}
\address[label2]{Dept. of Computer Science and Artificial Intelligence, DaSCI, University of Granada, 18071 Granada, Spain}
\address[label3]{Dept. of Botany, Faculty of Science, University of Granada, 18071 Granada, Spain}
\address[label4]{Dept. of Ecology, Faculty of Science, University of Granada, 18071 Granada, Spain}
\address[label5]{Interuniversity Institute of Earth System Research in Andalusia, Andalusian Center for the Environment (IISTA-CEAMA), Granada, 18071, Spain}
\address[label6] {Pyrenean Institute of Ecology, Spanish National Research Council (IPE-CSIC), 50059 Zaragoza, Spain}

\begin{abstract}
Monitoring the distribution and size of long-living large shrubs, such as junipers, is crucial for assessing the long-term impacts of global change on high-mountain ecosystems. While deep learning models have shown remarkable success in object segmentation, adapting these models to detect shrub species with polymorphic nature remains challenging. In this research, we release a large dataset of individual shrub delineations on freely available satellite imagery and use an instance segmentation model to map all junipers over the treeline for an entire biosphere reserve (Sierra Nevada, Spain). To optimize performance, we introduced a novel dual data construction approach: using photo-interpreted (PI) data for model development and fieldwork (FW) data for validation.  To account for the polymorphic nature of junipers during model evaluation, we developed a soft version of the Intersection over Union metric. Finally, we assessed the uncertainty of the resulting map in terms of canopy cover and density of shrubs per size class. Our model achieved an F1-score in shrub delineation of 87.87\% on the PI data and 76.86\% on the FW data. The R2 and RMSE of the observed versus predicted relationship were 0.63 and 6.67\% for canopy cover, and 0.90 and 20.62 for shrub density. The greater density of larger shrubs in lower altitudes and smaller shrubs in higher altitudes observed in the model outputs was also present in the PI and FW data, suggesting an altitudinal uplift in the optimal performance of the species. This study demonstrates that deep learning applied on freely available high-resolution satellite imagery is useful to detect medium to large shrubs of high ecological value at the regional scale, which could be expanded to other high-mountains worldwide and to historical and forthcoming imagery.
\end{abstract}


\end{frontmatter}

\pagebreak

\section{Introduction} \label{S1}
Climate change is forcing species to move latitudinally and altitudinally to maintain their climatic optimum. Species are often moving rapidly over large geographic areas, so methodological tools are needed to track these massive movements and to identify the dynamics of advancing and retreating fronts. 

The study of the distribution and abundance of organisms has always been a fundamental tenet in ecological science \citep{krebs2013ecology}. Classical field surveys (e.g., transects, plots) to count individuals are the most accurate, but it is impractical to use them when trying to identify and count individuals over vast  and frequently remote areas. New methodological tools need to be developed to enable ecologists to identify, count and map individuals over large areas with precision.

High-mountain shrubs play a vital role in ecosystems, contributing significantly to soil stabilization in the headwaters of watersheds, carbon sequestration, wildlife habitat provision, microclimate moderation, and overall biodiversity support \citep{adhikari2017aboveground}.  Climate change may intensify the vulnerability of these species, and reshape their geographical ranges to more climatically suitable regions
\citep{el2023potential}.
In this context, the development of high-precision maps at individual level is indispensable for an  accurate, but efficient and timely tracking of shrub distribution \citep{otto2012reconciling}. This precise but quick mapping is essential for various purposes, including environmental monitoring, biodiversity conservation, forestry, climate impact assessment, invasive species monitoring, land management, and urban planning \citep{ayhan2020vegetation}.

Combining remote sensing (RS) and artificial intelligence (AI) technologies can provide a great opportunity to improve \textit{in-situ} field surveying by opening up opportunities for automation. RS technologies offer highly detailed spatial resolution granting exceptional flexibility in data acquisition. This data can be afterwards processed by deep learning (DL) models for automatic identification of shrubs.  



Many studies have used remote sensing data to generate land cover maps including shrublands \citep{soubry2022mapping}. These studies capture a broad distribution of shrubs without delineating them individually.
Only a few attempts have been made to identify a specific types of shrubs (Table \ref{tab1}).
These studies are constrained by the following limitations. (1) Most of them were conducted across three distinct ecosystems. Mountain areas have been tackled once. (2) Only one study used satellite data \citep{guirado2021mask}.
(3) The studied shrubs have a consistent morphology. (4) None of them deployed the model to generate a large scale distribution of shrubs.

\begin{sidewaystable}
\centering
\caption{Review of studies using remote sensing and machine/deep learning to identify shrub individuals.}
\label{tab1}
\resizebox{\textwidth}{!}{%
\begin{tabular}{cccccccccccc}
\hline
\textbf{\thead{Shrub species}} & \textbf{\thead{RS \\ technology}} & 
\textbf{\thead{Spatial\\ resolution}} & \textbf{\thead{Data type}} & 
\textbf{\thead{Data size}} & 
\textbf{\thead{publicly \\available}} & \textbf{\thead{Mapping \\ approach}} & \textbf{\thead{Model}} & 
\textbf{\thead{Study area}} & 
\textbf{\thead{Ecosystem type}} & \textbf{\thead{Reference}} \\

\hline

\textit{\thead{Hakea suaveolens\\ and \\Hakea sericea}} & \thead{UAV} & \thead{20 cm} & \thead{RGB} & \thead{100 images \\with shrubs} & \thead{No} & \thead{Semantic\\ segmentation} & \thead{U-Net} & \thead{South Africa} & \thead{Fynbos} & \thead{\cite{james2020detecting}} \\

\hline

\thead{4 species} & \thead{UAV} & \thead{1.5 cm} & \thead{RGB} & \thead{2991 shrubs} & \thead{No} & \thead{Semantic\\ segmentation} & \thead{OBIA + \\Random Forest} & \thead{Northwest\\ China} & \thead{Dryland} & \thead{\cite{li2021classifying}} \\

\hline

\textit{\thead{Hakea sericea}} & \thead{UAV} & - & \thead{RGB} & \thead{100 images\\ with shrubs} & \thead{No} & \thead{Patch-based \\ classification and   \\ Semantic segmentation} & \thead{Xception\\ Inception \\ MobileNet \\ U-Net} & \thead{South Africa} & \thead{Fynbos} & \thead{\cite{james2021shrub}} \\

\hline

\textit{\thead{pearl bluebush \\ and \\ Maireana sedifolia}} & \thead{UAV} & \thead{(0.8, 2, 3) cm} & \thead{RGB} & \thead{4111 shrubs} & \thead{No} & \thead{Object detection} & \thead{Faster-RCNN \\ YOLO-V3 \\ SSD} & \thead{South of Australia} & \thead{Dryland} & \thead{\cite{retallack2022using}} \\

\hline

\textit{\thead{Cytisus scoparius}} & \thead{LiDAR} & \thead{2 cm} & \thead{Structural\\ features} & \thead{65 shrubs} & \thead{No} & \thead{3D point \\detection} & \thead{Machine learning \\models} & \thead{Denmark} & \thead{Grassland} & \thead{\cite{madsen2020detecting}} \\

\hline

\textit{\thead{Ziziphus lotus\\ (scattered shrubs)}} & \thead{GE\\ satellite} & \thead{12 cm} & \thead{RGB} & \thead{82 shrubs} & \thead{No} & \thead{Patch-based\\ classification} & \thead{GoogleNet\\ and \\ResNet} & \thead{South of Spain \\and \\Cyprus} & \thead{Dryland} & \thead{\cite{guirado2017deep}} \\

\hline

\textit{\thead{Ziziphus lotus\\ (scattered shrubs)}} & \thead{GE\\ satellite\\ and UAV} & \thead{1m} & \thead{RGB} & - & \thead{Under request} & \thead{Instance\\ segmentation} & \thead{Mask R-CNN \\ + OBIA} & \thead{South of Spain} & \thead{Dryland} & \thead{\cite{guirado2021mask}} \\

\hline

\thead{6 species} & \thead{UAV} & \thead{1.5 - 2.6 cm} & \thead{RGB} & \thead{-} & \thead{Yes} & \thead{Image\\ classification} & \thead{ConvNeXt\\Resnet \\Swin \\ Vit} & \thead{Northeast of Japan} & \thead{High-mountain} & \thead{\cite{moritake2024sub}} \\

\hline

\textit{\thead{Juniperus}} & \thead{GE \\Satellite} & \thead{13 cm} & \thead{RGB} & \thead{8580 shrubs} & \thead{Yes} & \thead{Instance\\ segmentation} & \thead{Mask R-CNN} & \thead{South of Spain} & \thead{High-mountain} & \thead{\textbf{Ours}} \\

\hline

\end{tabular}%
}
\end{sidewaystable}

The delineation of plant species featuring morphological variations from satellite and even from aerial imagery has been often addressed in ecological and remote sensing applications, but it continues to be challenging \citep{ramirez2022quantitative}.
In general, shrub delineation using Convolution Neural Networks (CNNs) can be effective when these shrubs have consistent patterns. However, this process may become more challenging in highly diverse ecosystems \citep{zhang2020identifying} and in scenarios with polymorphic shrubs, which are prone to overlapping and splitting due to canopy thinning \citep{dong2019progressive}. 
These scenarios make the data annotation process challenging and uncertain.

The research reported in this paper aims to demonstrate that freely available RGB high-resolution satellite imagery are useful to detect medium to large shrubs of high ecological value on a regional scale across large areas.
Our main objectives can be summarized as follows:

\begin{itemize}

\item We proposed the largest publicly available dataset of polymorphic shrubs ($8,580$ digitized individuals).

\item We introduced a new data construction approach to overcome the limitations of field surveying methods.

\item We developed a soft version of the Intersection over Union (IoU) metric for model evaluation.

\item We deployed the model at the large scale and generated a map of high-mountain shrubs, then analyzed their distribution.

\end{itemize}

The structure of this study is organized as follows: Section \ref{S2} describes the study area and materials. Section \ref{S3} outlines the methodology. Section \ref{S4} presents the obtained results. Section \ref{S5} presents the discussion. Section \ref{S6} summarizes the key findings of the study and sheds light onto future works.

\section{Study area and materials} \label{S2}

\subsection{Study area}
This research occurred in the Sierra Nevada National Park located on the southern fringe of Iberia at $37^o 06'$N $3^o05'$  W, between the provinces of Granada and Almería, Spain (Fig. \ref{study_area}-b). This park contains the highest mountains in western Europe after the Alps, with an elevation of 3479 m at the Mulhacén peak \citep{palacios2020climate}. This ecosystem includes an abundance of two long-living shrubs, named \textit{Juniperus communis} and \textit{Juniperus sabina}, with priority conservation interest at the European level.

\subsection{Target species}
\begin{figure}[H]
  \begin{center}
    \includegraphics[width=10cm]{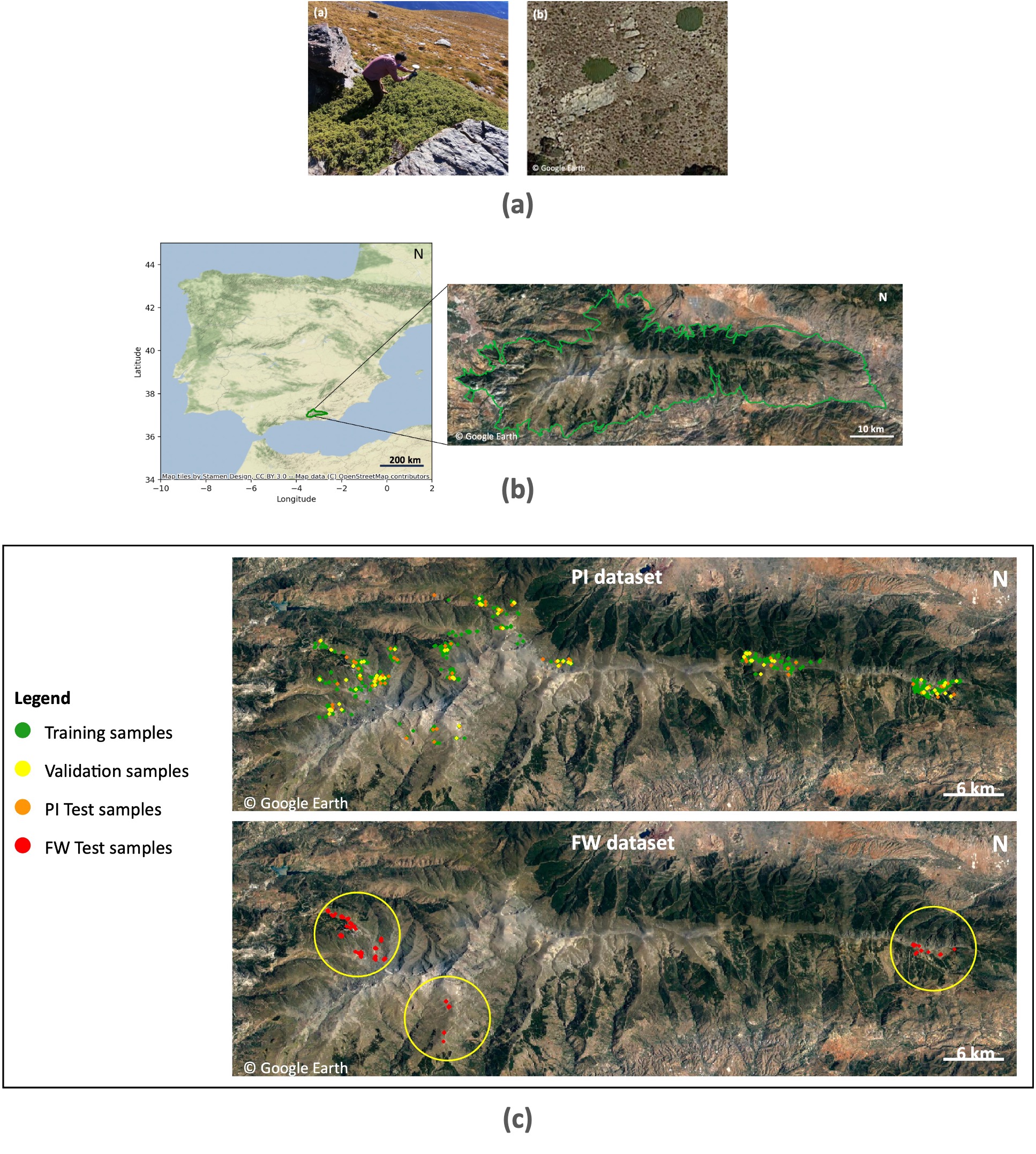} 
  \end{center}
  \caption{(a) juniper shrub captured in-situ(left) and from Google Earth satellites (right). (b) Location of the study area in the National Park of Sierra Nevada (Andalusia, Spain). (c) Distribution of Photo Interpreted (PI) and Field Work (FW) datasets.}
  \label{study_area}
\end{figure}

Common juniper, \textit{Juniperus communis} L. (\textit{Cupresaceae}), is among the most widely distributed gymnosperms in the Holarctic, ranging from circum-Mediterranean mountains up to subarctic tundra \citep{garcia2000geographical}. 
Juniper (Fig. \ref{study_area}-a) is a typical dwarf evergreen needle-leaf long-living shrub that occurs on poor soils and harsh environments. This species shows a continuous distribution in northern and central Europe, but populations become progressively more fragmented towards the Mediterranean Basin, where the species is located exclusively in high-mountain areas, dominating the strip between the tree-line and the woody-line. 
These populations, such as those in the southern Iberian Peninsula, are characterized by a very low regeneration ability under natural conditions.
There, populations are currently dominated by adult and senescent individuals, with extremely low proportions of seedlings and juveniles \citep{garcia1999age}.



\begin{figure}[H]
  \begin{center}
    \includegraphics[width=11cm]{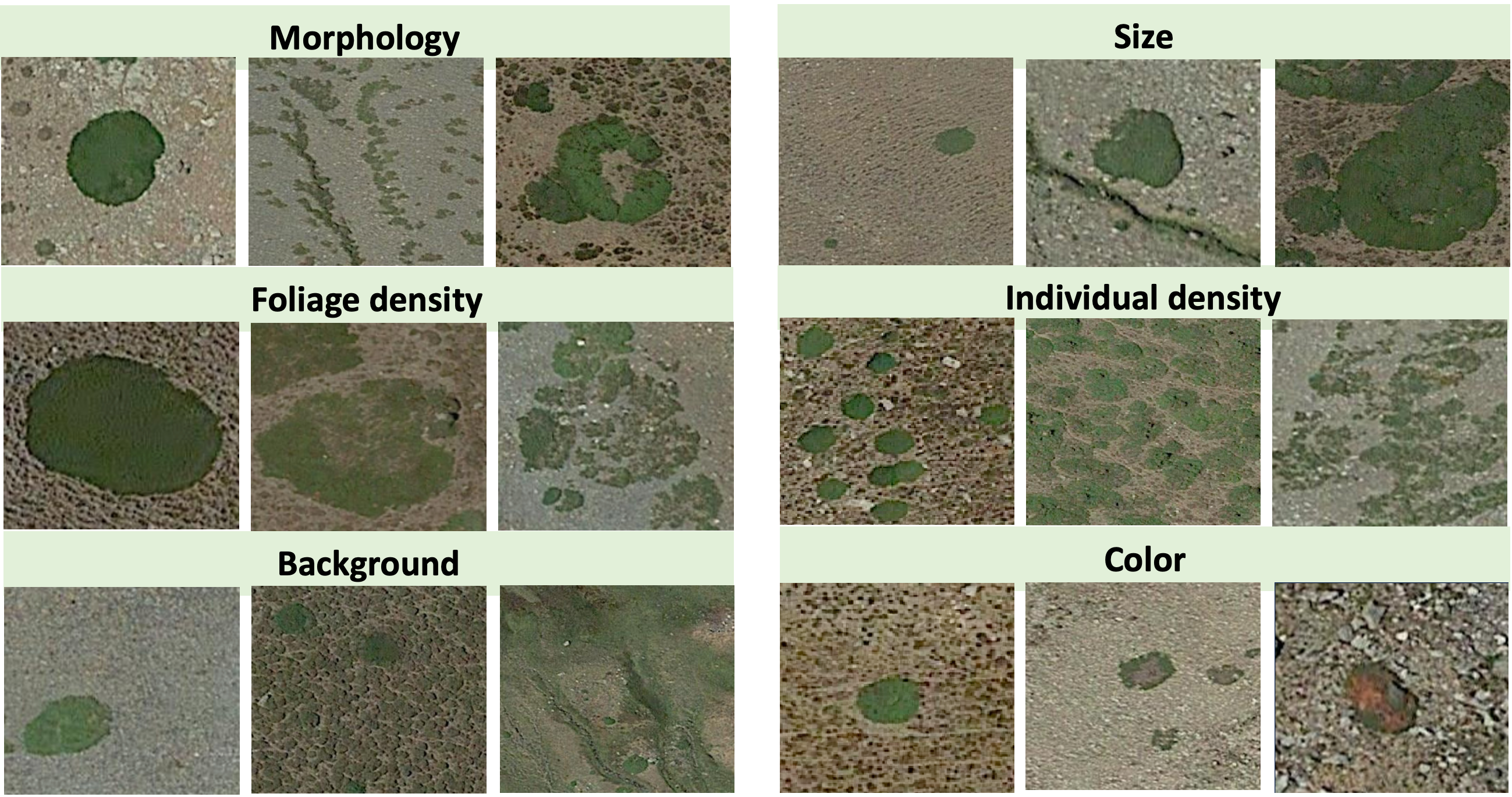} 
  \end{center}
  \caption{Variation in the growing pattern of juniper shrubs.}
  \label{shrubs}
\end{figure}

In the complex terrain of the Sierra Nevada high mountains, the highly heterogeneous backgrounds present significant challenges. These difficulties are further compounded by the polymorphic nature of juniper shrubs, whose crowns can vary greatly between similar individuals as follows (Fig. \ref{shrubs}):
(1) They can grow in different morphologies (e.g., hemispherical, stripes, crescent shape, thinned lines, etc.) depending on their age, slope, and altitude of their geographic location. 
(2) They can grow in different individual densities: isolated or in colonies that merge to form big shrubs. This latter pattern makes their individual detection either by human experts or by the model itself particularly challenging. This is even truly challenging during the fieldwork surveys to collect ground truth validation samples.
(3) They can grow in different sizes, ranging from centimeters to hundreds of meters which sometimes can cover the entire image tile fed to the model. This makes the model unable to see the full object and can confuse it with lakes and grasslands. 
(4) They can have different colors. They are not always green but they can be brown, red, and with different green shades depending on their health, season, and time of the day. This makes the foreground/background problem even harder.
(5) They can have different foliage density (i.e., leaf area index) and crown vertical patterns depending on the age and health of the shrub.

\subsection{Data acquisition}
Training DL models requires large and high-quality datasets. However, collecting such data through fieldwork is expensive, labor-intensive, time-consuming, unsustainable, and limited to a small spatial scale. To address these challenges, we propose the combination of a large amount of less accurate but easier to get photo-interpreted (PI) data, with a smaller amount of high quality but costly to get field-work (FW) data.

The PI data was annotated by botanists who visually inspected a representative range of sites in the satellite images to annotate all juniper shrubs in each site, without conducting field visits. As a result, this data may contain some uncertainty and errors in the annotations, as identifying junipers can be challenging with the 13-cm resolution of the satellite images. The PI data contains samples of 712 sites of 448x448m with a total of 6809 juniper shrubs. The FW data was obtained in the field by botanists and ecologists with a differential centimetric GPS. A total of 124 sample sites of 420x336m with a total of 1771 juniper shrubs were visited. Every shrub present in each site was georeferenced. The stored coordinates were then overlapped onto the satellite image for adjustment, verification and clipping. The sites for collecting the FW data were different from the sites where the PI data was extracted. The average minimum distance between the FW sites and the PI sites was 433 m (Table \ref{tab3} and \ref{tab2})).

\begin{table}[H]
\centering
\caption{Distribution of the number of images and instances in PI and FW datasets over different shrub sizes.}
\label{tab3}
\resizebox{\textwidth}{!}{%
\begin{tabular}{clcrrrrrrrr}
\hline
\multirow{2}{*}{\textbf{\thead{Data name}}} & \multicolumn{1}{c}{\multirow{2}{*}{\textbf{\thead{Data \\ partitions}}}} & \multirow{2}{*}{\textbf{\thead{Size \\ of images}}} & \multicolumn{1}{c}{\multirow{2}{*}{\textbf{\thead{Number \\ of images}}}} & \multicolumn{7}{c}{\textbf{\thead{Number of shrubs}}} \\
 & \multicolumn{1}{c}{} &  & \multicolumn{1}{c}{} & \multicolumn{1}{c}{\textbf{\thead{All}}} & \multicolumn{1}{c}{\textbf{\thead{XS}}} & \multicolumn{1}{c}{\textbf{\thead{S}}} & \multicolumn{1}{c}{\textbf{\thead{M}}} & \multicolumn{1}{c}{\textbf{\thead{L}}} & \multicolumn{1}{c}{\textbf{\thead{XL}}} & \multicolumn{1}{c}{\textbf{\thead{XXL}}} \\
 \hline
\multirow{3}{*}{Photo Interpreted (PI)} & Train & \multirow{3}{*}{\thead{(448 x 448) pixels \\ 3582.33 $m^2$}} & 570 & 5459 & 559 & 837 & 1392 & 1331 & 805 & 535\\
 & Validation &  & 67 & 660 & 90 & 91 & 152 & 170 & 97 & 60\\ 
  & PI Test &  & 75 & 690 & 29 & 91 & 163 & 201 & 119 & 87\\
  \hline
\multicolumn{1}{l}{Field Work (FW)} & FW Test & \thead{(420 x 336) pixels\\ 2550.68 $m^2$} & 124 & 1771 & 310 & 329 & 439 & 351 & 194 & 148 \\
\hline
\multicolumn{3}{c}{Total} & 836 & 8580 & 988 & 1348 & 2146 & 2053 & 1215 & 830 \\
\hline
\end{tabular}%
}
\end{table}

One of the main goals of our study is to create a model that can produce a precise and accurate map of junipers useful for ecologists and managers. This requires that the model generalizes well across the different morphologies of junipers, which usually relate to the different environmental conditions of the high-mountain ecosystems. For these reasons, when constructing the PI and FW datasets, we considered the following constraints: (1) High variation in growing patterns: junipers exhibit significant variation in morphology, density, size, canopy/foliage, and color within and across different areas in the park. This variation necessitates careful dataset construction to ensure that the model can generalize well to all types of junipers. (2): High ecosystem diversity of Sierra Nevada National park: Sierra Nevada embraces one of the most diverse environments in Europe and in the Mediterranean Basin \citep{canadas2014hotspots}. Such environmental heterogeneity partially determines the differences in morphology of junipers across the different environments and provides different backgrounds where the species grows, from homogeneous barren lands and meadows, to heterogeneous shrublands and abandoned croplands, where identifying junipers presents a more challenging foreground/background problem. 

Thus, to create PI and FW datasets, we employed a block splitting approach \citep{roberts2017cross, uieda2018verde}. Using expert knowledge, we identified various sites across the park where junipers are likely to grow. These sites were divided into a grid of patches. The experts involved in this work constructed PI dataset by visually filtering out all patches that lacked junipers and those where visual inspection of junipers was challenging. From the remaining patches, we randomly sampled the PI subsets (80\% of training to get the optimal parameters of the model, 10\% of validation to avoid overfitting and select the best model configuration, and 10\% of PI test to evaluate the generalization of the model on the PI dataset) ensuring that their patches were spatially distant from each other to prevent adjacent patches from being used for both training and testing. Within each patch, all the existing junipers were annotated (Fig. \ref{study_area}-c).

For the FW dataset, we utilized three different accessible sites. From these sites, we manually selected patches that were spatially distant from the PI patches and situated on reasonably accessible slopes, allowing experts to easily conduct field visits. This dataset contains only one subset that we call FW Test set used for two key purposes: (1) evaluate the generalization performance of the model on juniper samples from the FW data, and (2) provide an estimate of performance for the wall-to-wall juniper map of Sierra Nevada.

\section{Methods} \label{S3}

\subsection{Study design}
\begin{figure}[H]
  \begin{center}
    \includegraphics[width=12cm]{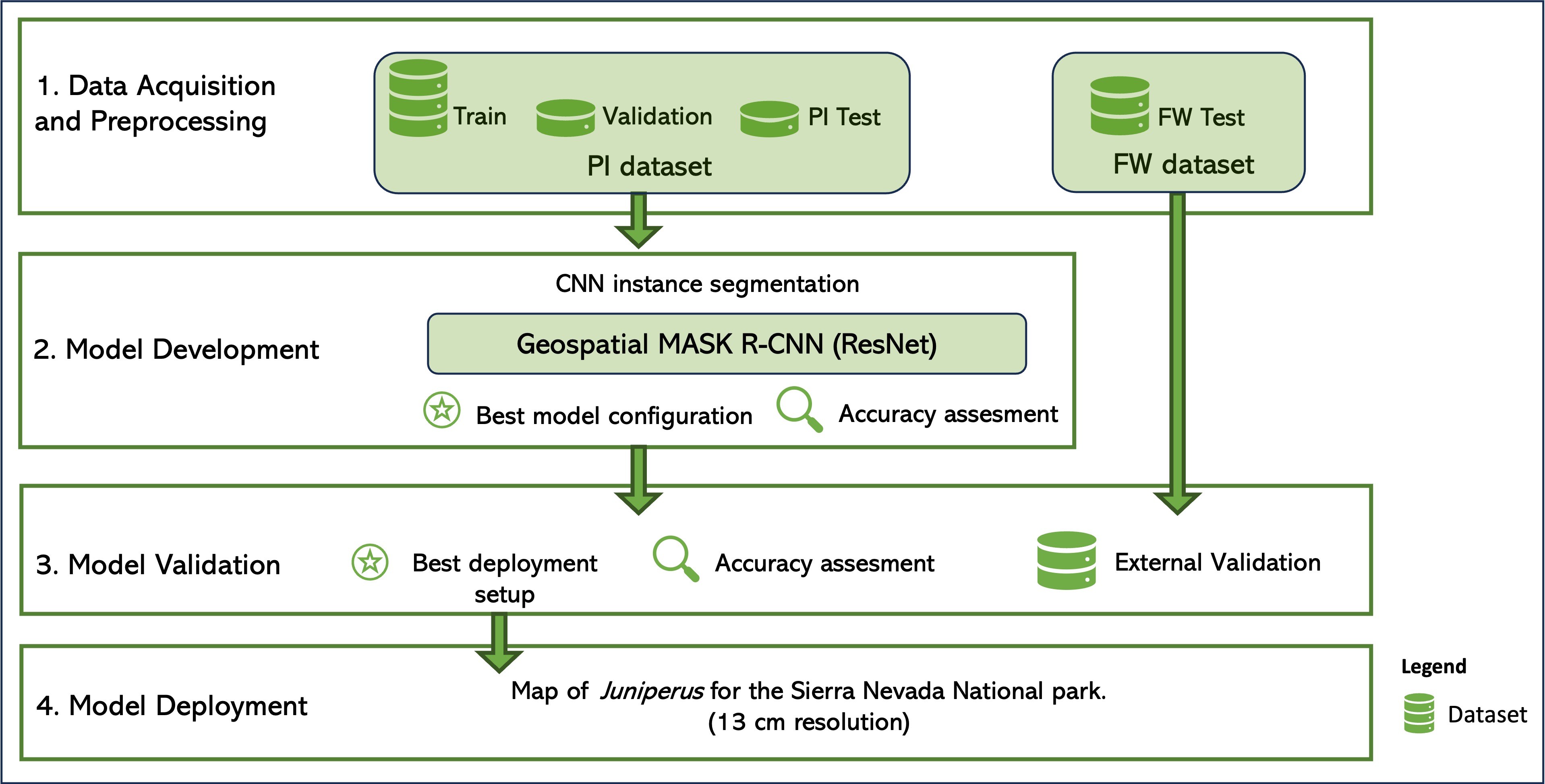} 
  \end{center}
  \caption{Workflow of the automatic delineation of juniper shrubs.}
  \label{Workflow}
\end{figure}

The design of this study is organized into four main steps (Fig. \ref{Workflow}): (1) the PI and the FW data were collected, annotated,  and preprocessed, (2) the shrub delineation model was developed using the PI data, then (3) the developed model was validated using the FW data, finally (4) the model was deployed to generate a wall-to-wall map of juniper. 

\begin{figure}[H]
  \begin{center}
    \includegraphics[width=12cm]{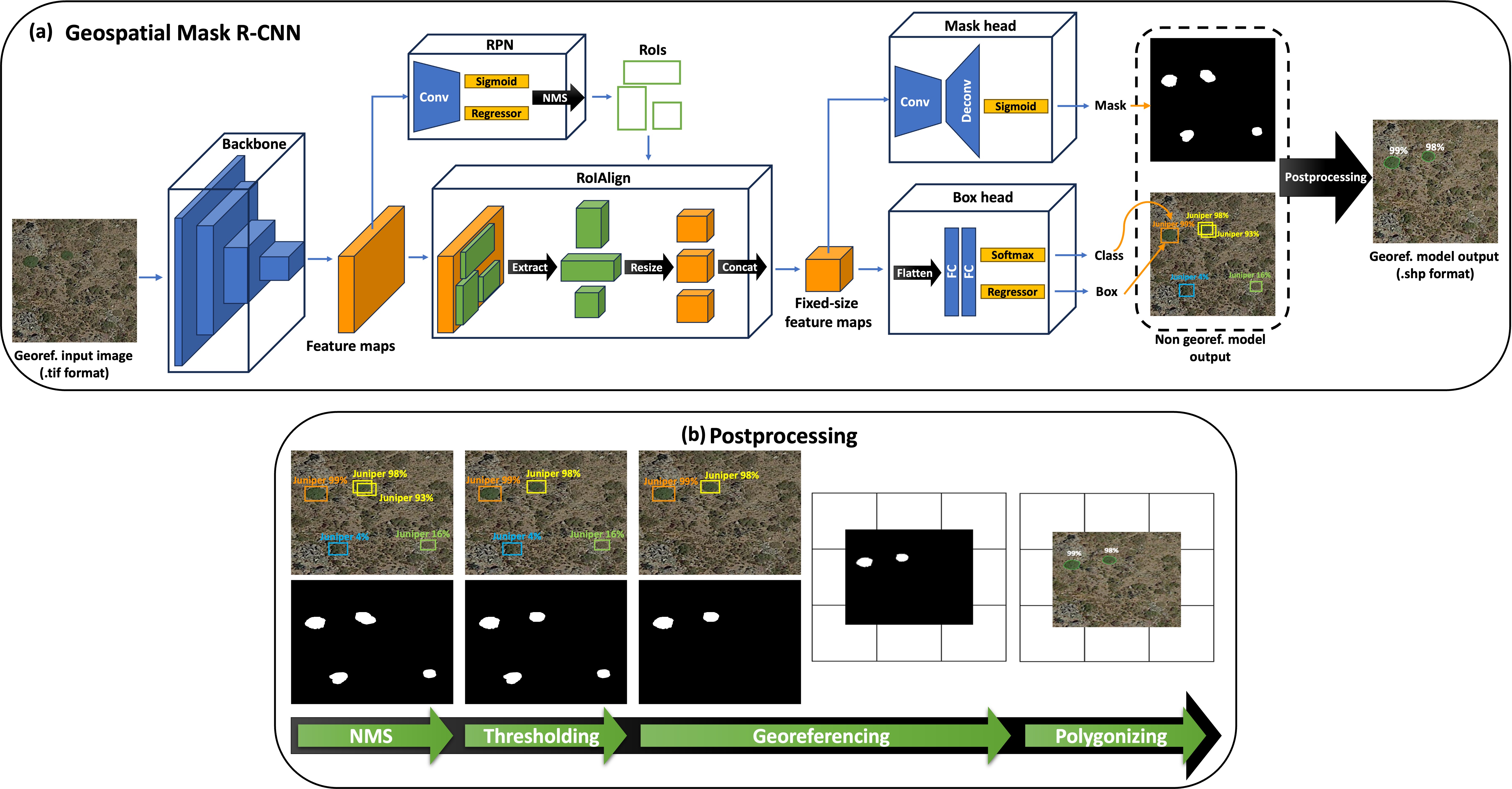} 
  \end{center}
  \caption{Description of the model design used to delineate juniper shrubs.}
  \label{MASK_RCNN}
\end{figure}

Mask-RCNN model \citep{he2017mask} has shown impressive results in a variety of vegetation detection applications \citep{zheng2022deep, kierdorf2023growliflower}. In the context of our study, We finetuned and optimized different architectures proposed by Detectron2 library \citep{wu2019detectron2}. Fig. \ref{MASK_RCNN} presents the model design. An instance segmentation-based approach was employed to individually delineate junipers, enabling more effective monitoring of their changes over time. 
During the training, the model was evaluated on the validation set after each 10 epochs based on which the best model state was saved.

\subsection{Model evaluation}

To evaluate instance segmentation-based DL models, we usually use the Intersection over Union (IoU) metric. The evaluation process includes two main steps: (1) predictions' evaluation where the algorithm iterates over the predicted instances to compute the number of TPs (True Positives) and FPs (False Positives), and (2) labels' evaluation where the algorithm iterates over the ground truth instances to get the number of FNs (False Negatives). 
This metric computes the area of the intersection between the prediction $p$ and the label (i.e., ground truth) $l$ and divide the result by the area of their union (Eq. \ref{eq1}). It always selects the best matching label or prediction in the evaluation process.

\begin{equation}
    \text{IoU}(p, l) = \frac{Area(p \cap l)}{Area(p \cup l)}
    \label{eq1}
\end{equation}

\begin{figure}[H]
  \begin{center}
    \includegraphics[width=8cm]{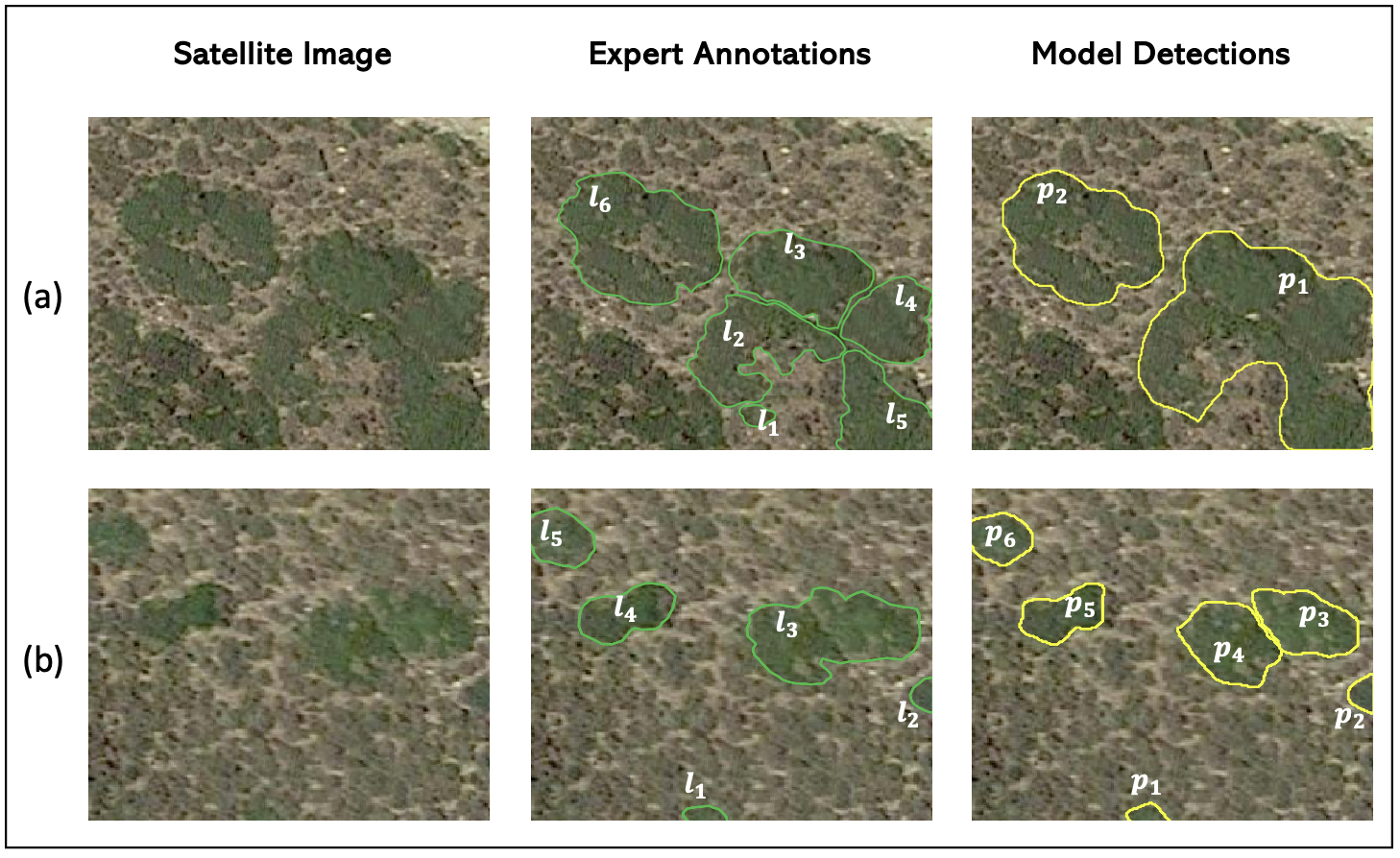} 
  \end{center}
  \caption{Two examples (a) and (b)  showing a discrepancy between expert annotations and model detections due to the polymorphic character of junipers.}
  \label{IoU_limits}
\end{figure}

Due to the polymorphic nature of junipers, their identification and annotation by experts becomes challenging. One expert may identify a shrub as one individual while others may see it as multiple individuals. As a result, the model may exhibit similar behavior in its predictions which can be strongly penalized by the IoU metric (Fig. \ref{IoU_limits}). From the ecological perspective, detecting one juniper as multiple instances and vice versa are acceptable due to the complex nature of these species.

Thus, to evaluate the performance of DL models with respect to polymorphic plant species subject to overlapping, splitting due to canopy thinning, and uncertain human experts annotations, we developed a soft version of IoU that we name S-IoU. The two metrics convey different but complementary information:
(1) The IoU evaluates how precise the model is in detecting junipers while being aligned with human experts annotations. (2) The S-IoU evaluates the proportion of junipers’ areas being detected by the model since we divide the intersection by the ground truth area and we use all matching ground truth shrubs instead of the best matching shrub. Unlike the IoU that evaluates the predictions and labels similarly, this metric evaluates them differently: Eq. \ref{eq2} was used to identify the number of TPs and FPs, while Eq. \ref{eq3} was used to identify the number of FNs (Fig. \ref{IOU_vs_MIOGTA}).

\begin{equation}
    \text{S-IoU}(p, S_{l_{match}}) = \frac{Area(p \cap \bigcup_{k=1}^{|S_{l_{match}}|} l_k)}{Area(\bigcup_{k=1}^{|S_{l_{match}}|} l_k)}
    \label{eq2}
\end{equation}

\begin{equation}
    \text{S-IoU}(l, S_{p_{match}}) = \frac{Area(l \cap \bigcup_{k=1}^{|S_{p_{match}}|} p_k)}{Area(l)}
    \label{eq3}
\end{equation}

Where:
$S_{p_{match}}$ the set of all matching predictions. $S_{l_{match}}$ is the set of all matching labels. $l_k$ and $p_k$ are the label and prediction of id $k$, respectively.

To evaluate the overall model performance, three metrics were used: (1) the Precision (\ref{eq5}) to assess how precise the model is in delineating the shrubs, (2) the Recall (\ref{eq6}) to evaluate how accurate the model is in recalling the shrubs, and (3) the F1-score (\ref{eq7}) to examine the ability of the model to maintain the trade-off precision-recall.

\begin{equation}
    \text{Precision} = \frac{\text{TP}}{\text{TP+FP}}
    \label{eq5}
\end{equation}

\begin{equation}
    \text{Recall} = \frac{\text{TP}}{\text{TP+FN}}
    \label{eq6}
\end{equation}

\begin{equation}
    \text{F1-score} = \frac{2*\text{Precision}*\text{Recall}}{\text{Precision} + \text{Recall}}
    \label{eq7}
\end{equation}

\subsection{Model deployment}
\begin{figure}[H]
\centering
  \begin{center}
    \includegraphics[width=6cm]{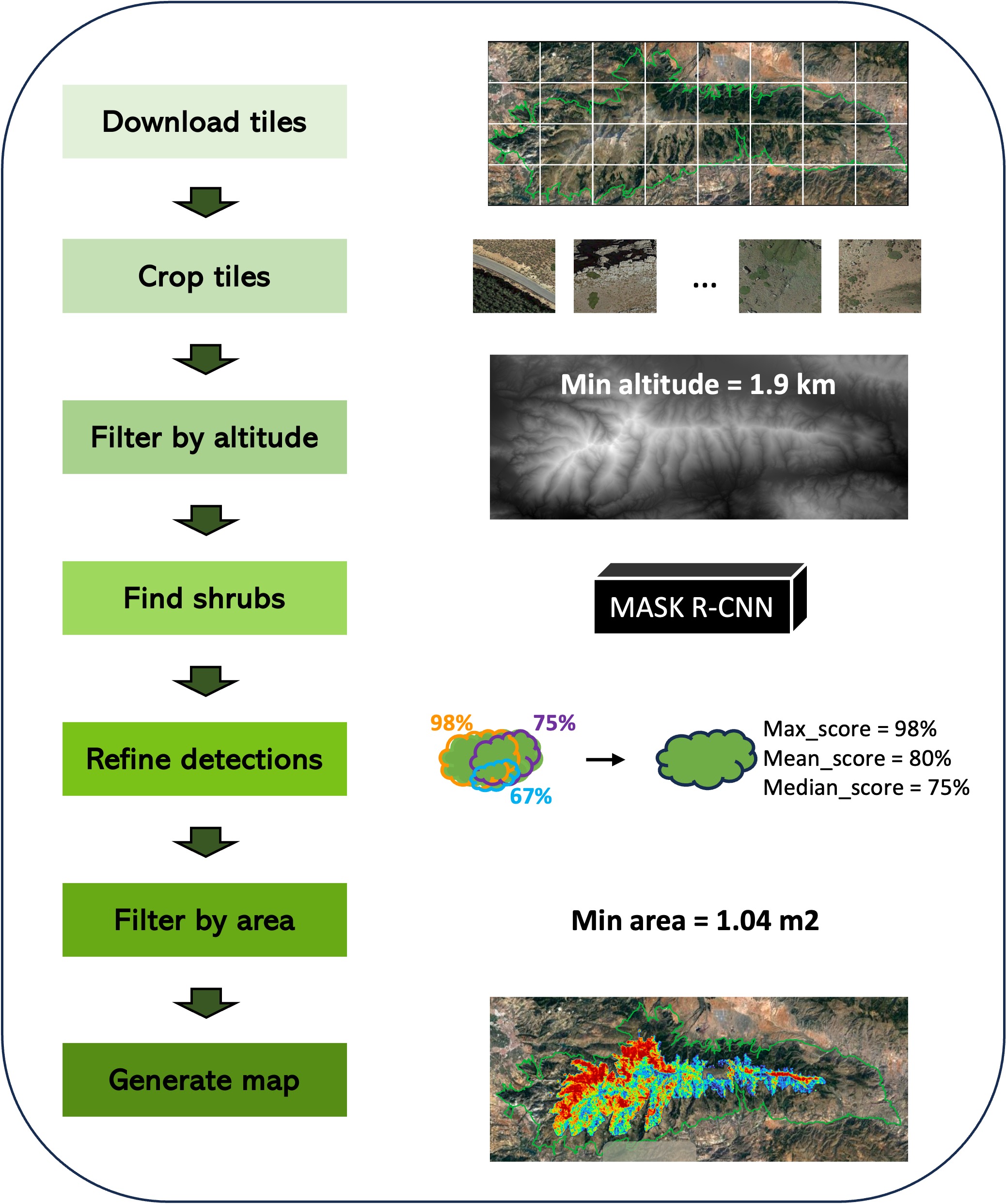} 
  \end{center}
  \caption{Description of the deployment pipeline.}
  \label{deploy}
\end{figure}

After developing the model using the PI data and validating it using the FW data, it was deployed over all the park to generate a wall-to-wall map of juniper. The deployment pipeline consists in seven main steps (Fig. \ref{deploy}):
(1) A set of RGB tiles covering the whole park were downloaded at 13 cm resolution. (2) Each tile was cropped into images of size (448 x 448). 
(3) A DEM (Digital Elevation Model) at 2m/pixel resolution was used to filter out images with altitude less than 1.9 km because we assume that juniper is more likely to grow above this altitude and the model is more prone to make false detections below this altitude. To perform this filtering, each image was assigned an altitude value corresponding to the maximum altitude covered by the image.
(4) The filtered images were fed to the model. 
(5) A refining process was applied to the model detections that consists in dissolving the multiple detections of the same shrub in one detection using the union over their geometries and creating three kinds of scores (average, median, and maximum scores).
(6) A second area-based filtering was applied, where the detections with area less than $1.04m^2$ (i.e., corresponding to the 10th percentile of the detected areas) were filtered out because we are less confident in the model detections below this threshold.
(7) All the detections of the images were merged to generate a final map of juniper at 13cm resolution.

\subsection{Statistical Analysis}
To assess the model's generalization performance, we evaluate it on both the PI and FW test sets. Additionally, we analyze the model's performance across six different shrub size ranges (Table \ref{tab9}). For deployment validation, we use the FW data to examine the correlation between the number of predicted and observed junipers in FW images, as well as the correlation between the proportion of the FW image occupied by predicted versus observed junipers. Furthermore, we investigate the distribution of junipers by size across different altitudes, comparing results from the PI data, FW data, and the model output trained on PI data.

\section{Results} \label{S4}
\subsection{Analysis of model evaluation results}
The optimal model configuration was achieved with the ResNet101-C4 backbone, a batch size of two, and an initial learning rate of 0.0025. Data augmentation was not utilized, as it did not yield a significant improvement in model performance. (Table \ref{tab5}).

\begin{figure}[H]
  \begin{center}
    \includegraphics[width=10cm]{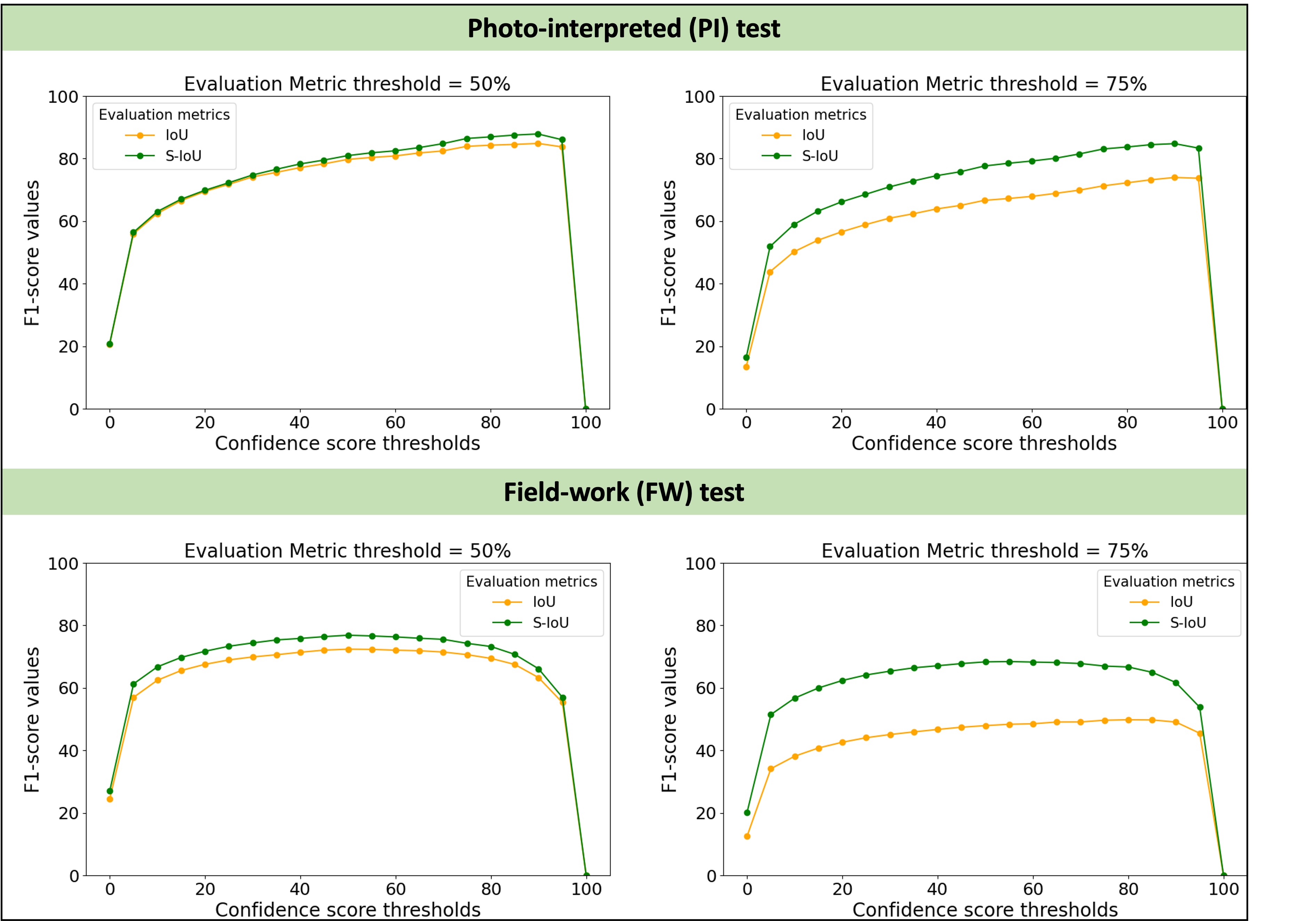} 
  \end{center}
  \caption{Mask R-CNN performance evaluated on PI and FW test sets using IoU and S-IoU metrics across different confidence score thresholds $\theta_{score}$.}
  \label{IOU_vs_MIOGTA_res}
\end{figure}

Figure \ref{IOU_vs_MIOGTA_res} illustrates the model's performance evaluated using IoU and S-IoU metrics across various confidence score thresholds ($\theta_{score}$). Both metrics follow a similar pattern, though the F1-scores associated with the S-IoU metric are consistently higher than those with IoU, owing to the softer nature of S-IoU. This suggests that even when the model's predictions do not perfectly align with expert annotations, it can still detect a substantial portion of the juniper area.

When comparing the model's performance on the PI and FW test sets, distinct patterns emerge between the two curves. The F1-score curve for the PI test set shows a steady increase, reaching its peak at a confidence score threshold of $\theta_{score}=90\%$. In contrast, the F1-score for the FW test set initially rises, then stabilizes between thresholds of 20\% and 85\%, with a peak at $\theta_{score}=50\%$. This indicates that, unlike the PI data, the model's performance on the FW data remains relatively consistent across a wide range of confidence score thresholds. Since the FW data is a snapshot of real world data, a confidence score threshold of $\theta_{score}=50\%$ is recommended for deployment.

Table \ref{tab7} presents the model evaluation results for the PI and FW test sets, using the IoU and S-IoU metrics at two overlapping thresholds, 50\% and 75\%. The table also includes the corresponding values for true positives (TPs), false positives (FPs), false negatives (FNs), precision, recall, and F1-score.
The results indicate that the model performed best at the 50\% threshold. For the PI test set, the model achieved an F1-score of 84.84\% using the IoU metric and 87.87\% using the S-IoU metric. For the FW test set, the F1-scores were 72.39\% and 76.86\%, respectively. When comparing the model's performance on the PI and FW test sets, we observe a decrease of approximately 12\% in F1-score using IoU and 11\% using S-IoU at the 50\% threshold.

\begin{table}[H]
\centering
\small 
\caption{Description of the best Mask R-CNN configuration.}
\label{tab5}
\begin{tabular}{ll}
\hline
\textbf{\thead{Hyperparameters}} & \textbf{\thead{Best values}} \\
\hline
Maximum number of iterations & 4000 \\
Optimization algorithms & Momentum \\
Batch size & 2 \\
Initial learning rate & 0.0025 \\
Learning rate scheduler & WarmupCosine (2000 iterations, factor of 10) \\
Data augmentation & No augmentation \\
Maximum number of boxes & 256 \\
Feature extractor (backbone) & ResNet101-C4 \\
\hline
\end{tabular}%
\end{table}

\begin{table}[H]
\centering
\small
\caption{Mask R-CNN performance evaluated on PI test set (at $\theta_{score}=90\%$) and FW test set (at $\theta_{score}=50\%$) using IoU and S-IoU metrics at thresholds 50\% and 75\%.}
\label{tab7}
\begin{tabular}{cccrrrrrr}
\hline
\textbf{data name} & \textbf{metric name} & \textbf{metric threshold} & \multicolumn{1}{c}{\textbf{TP}} & \multicolumn{1}{c}{\textbf{FP}} & \multicolumn{1}{c}{\textbf{FN}} & \multicolumn{1}{c}{\textbf{Precision}} & \multicolumn{1}{c}{\textbf{Recall}} & \multicolumn{1}{c}{\textbf{F1-score}} \\ \hline
\multirow{4}{*}{PI Test} & \multirow{2}{*}{IoU} & 50 & 568 & 78 & 125 & 87.93 & 81.96 & 84.84 \\
& & 75 & 494 & 152 & 196 & 76.47 & 71.59 & 73.95 \\
& \multirow{2}{*}{S-IoU} & 50 & 572 & 74 & 84 & 88.55 & 87.20 & 87.87 \\ 
& & 75 & 551 & 95 & 103 & 85.29 & 84.25 & 84.77 \\
\hline
\multirow{4}{*}{FW Test} & \multirow{2}{*}{IoU} & 50 & 1268 & 438 & 529 & 74.32 & 70.56 & 72.39 \\
& & 75 & 833 & 873 & 938 & 48.82 & 47.03 & 47.91 \\
& \multirow{2}{*}{S-IoU} & 50 & 1307 & 399 & 388 & 76.61 & 77.11 & 76.86 \\
& & 75 & 1141 & 565 & 493 & 66.88 & 69.83 & 68.32 \\
\hline
\end{tabular}
\end{table}

\subsection{Assessment of shrub size on model performance}
Table \ref{tab9} shows the model evaluation results based on shrub size for both the PI and FW test sets, using the IoU and S-IoU metrics.

For the PI test set, the model performs well on medium (M), large (L), extra large (XL), and extra extra large (XXL) shrubs, achieving F1-scores above 81\% with IoU and above 83\% with S-IoU. However, for small (S) shrubs, the performance declines to 75\% with IoU and 78.53\% with S-IoU. The detection accuracy drops significantly for extra small (XS) shrubs, with F1-scores of 48.15\% and 53.06\% for IoU and S-IoU, respectively.

In the FW test set, the model achieves F1-scores greater than 82\% with IoU and 85\% with S-IoU for L and XL shrubs. For M and S shrubs, the F1-scores are above 70\% with IoU and 73\% with S-IoU. The performance declines for XS shrubs, with F1-scores of 54.09\% using IoU and 61.45\% using S-IoU. Interestingly, for XXL shrubs, the IoU-based F1-score is 72.86\%, while the S-IoU-based F1-score is 83.02\%. This indicates that while the model's detections for large shrubs covering a significant portion of the image may not align precisely with expert annotations, they are still identified as a colony of individuals using S-IoU.

\begin{table}[H]
\centering
\small
\caption{Mask R-CNN performance evaluated on PI test set (at $\theta_{score}=90\%$) and FW test set (at $\theta_{score}=50\%$) over different sizes using IoU and S-IoU metrics at 50\% threshold.}
\label{tab9}
\begin{tabular}{ccrrrrrr}
\hline
\multicolumn{1}{c}{\multirow{2}{*}{Data name}} &
  \multicolumn{1}{c}{\multirow{2}{*}{Size}} &
  \multicolumn{3}{c}{IoU} &
  \multicolumn{3}{c}{S-IoU} \\
\multicolumn{1}{c}{} &
  \multicolumn{1}{c}{} &
  \multicolumn{1}{c}{Precision} &
  \multicolumn{1}{c}{Recall} &
  \multicolumn{1}{c}{F1-score} &
  \multicolumn{1}{c}{Precision} &
  \multicolumn{1}{c}{Recall} &
  \multicolumn{1}{c}{F1-score} \\
\hline
\multirow{7}{*}{PI Test} 
 & XS & 54.17 & 43.33 & 48.15 & 54.17 & 52.00 & 53.06 \\
 & S & 78.75 & 71.59 & 75.00 & 80.00 & 77.11 & 78.53 \\
 & M & 84.77 & 78.05 & 81.27 & 84.77 & 82.05 & 83.39 \\
 & L & 95.03 & 88.21 & 91.94 & 94.48 & 92.94 & 93.70 \\
 & XL & 94.96 & 88.98 & 91.87 & 94.12 & 94.92 & 94.52 \\
 & XXL & 86.81 & 88.76 & 87.78 & 92.31 & 93.33 & 92.82 \\
 & All & 87.93 & 81.96 & 84.84 & 88.55 & 87.20 & 87.87 \\ \hline
 
\multirow{7}{*}{FW Test} 
 & XS & 58.24 & 50.50 & 54.09 & 63.22 & 59.78 & 61.45 \\
 & S & 71.39 & 68.70 & 70.02 & 73.19 & 74.09 & 73.64 \\
 & M & 70.31 & 73.09 & 71.67 & 71.21 & 78.77 & 74.80 \\
 & L & 84.14 & 81.59 & 82.85 & 84.42 & 86.88 & 85.63 \\
 & XL & 88.02 & 82.44 & 85.14 & 89.58 & 87.76 & 88.66 \\
 & XXL & 83.05 & 64.90 & 72.86 & 93.22 & 74.83 & 83.02 \\
 & All & 74.33 & 70.56 & 72.40 & 76.61 & 77.11 & 76.86 \\ \hline
\end{tabular}
\end{table}

\subsection{Analysis of model deployment results}
\begin{figure}[H]
\centering
  \begin{center}
    \includegraphics[width=13cm]{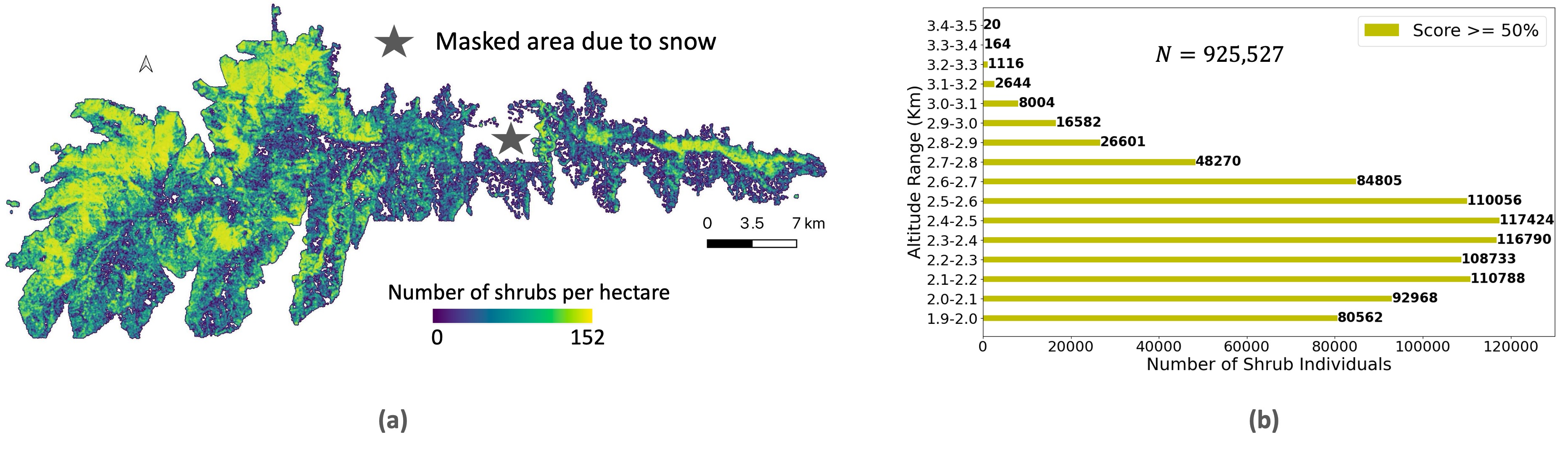} 
  \end{center}
  \caption{Density map highlighting (a) the spatial and the (b) altitudinal distribution of junipers, in the National Park of Sierra Nevada of Spain, at model score threshold $\theta_{score}=50\%$ using an altitude threshold of 1.9 km.}
  \label{map}
\end{figure}

\begin{figure}[H]
\centering
    \begin{center}
        \includegraphics[width=10cm]{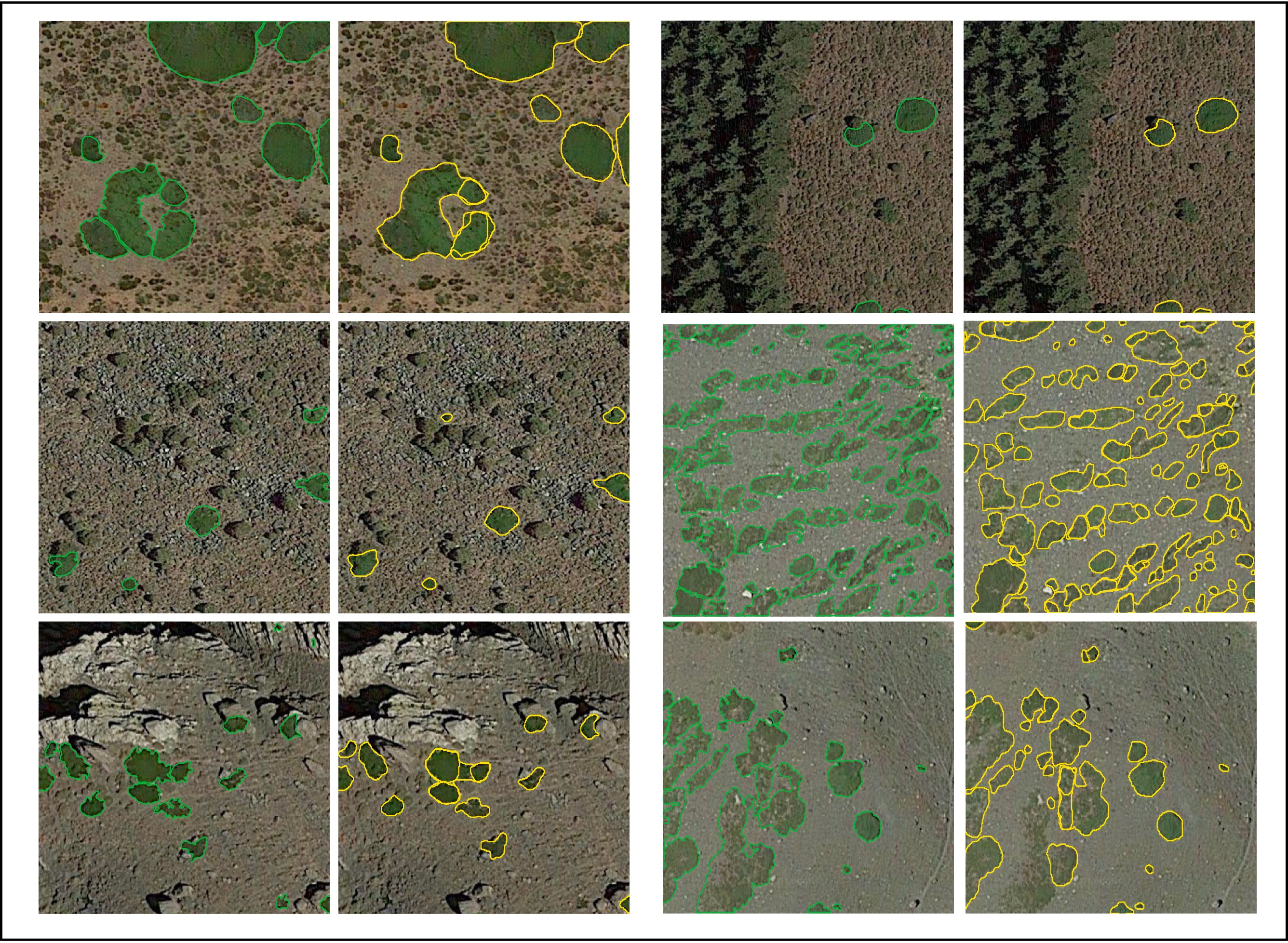}
    \end{center}
    \caption{Samples of model predictions (in yellow) and their corresponding experts annotations (in green).}
    \label{pred_samples}
\end{figure}

Fig \ref{map}-a displays the distribution of the detected junipers at model score threshold $\theta_{score}=50\%$. We can observe that junipers are highly concentrated in the North-West region and follow a stripe pattern in the North-East region, while they are less concentrated in the Southern region. The maximum number of junipers reach 152 individuals per hectare.
Fig. \ref{map}-b presents the distribution of juniper for each altitude range from 1.9km to 3.5km with a step of 100m at model score threshold $\theta_{score}=50\%$. 
Juniper individuals are highly concentrated within a specific range of altitude, mainly between 2km and 2.6km.
Fig. \ref{pred_samples} presents some samples of the model detections and their corresponding expert annotations at model confidence score threshold $\theta_{score}=50\%$.

\subsection{Validation of juniper map}
\begin{figure}[H]
\centering
  \begin{center}
    \includegraphics[width=14cm]{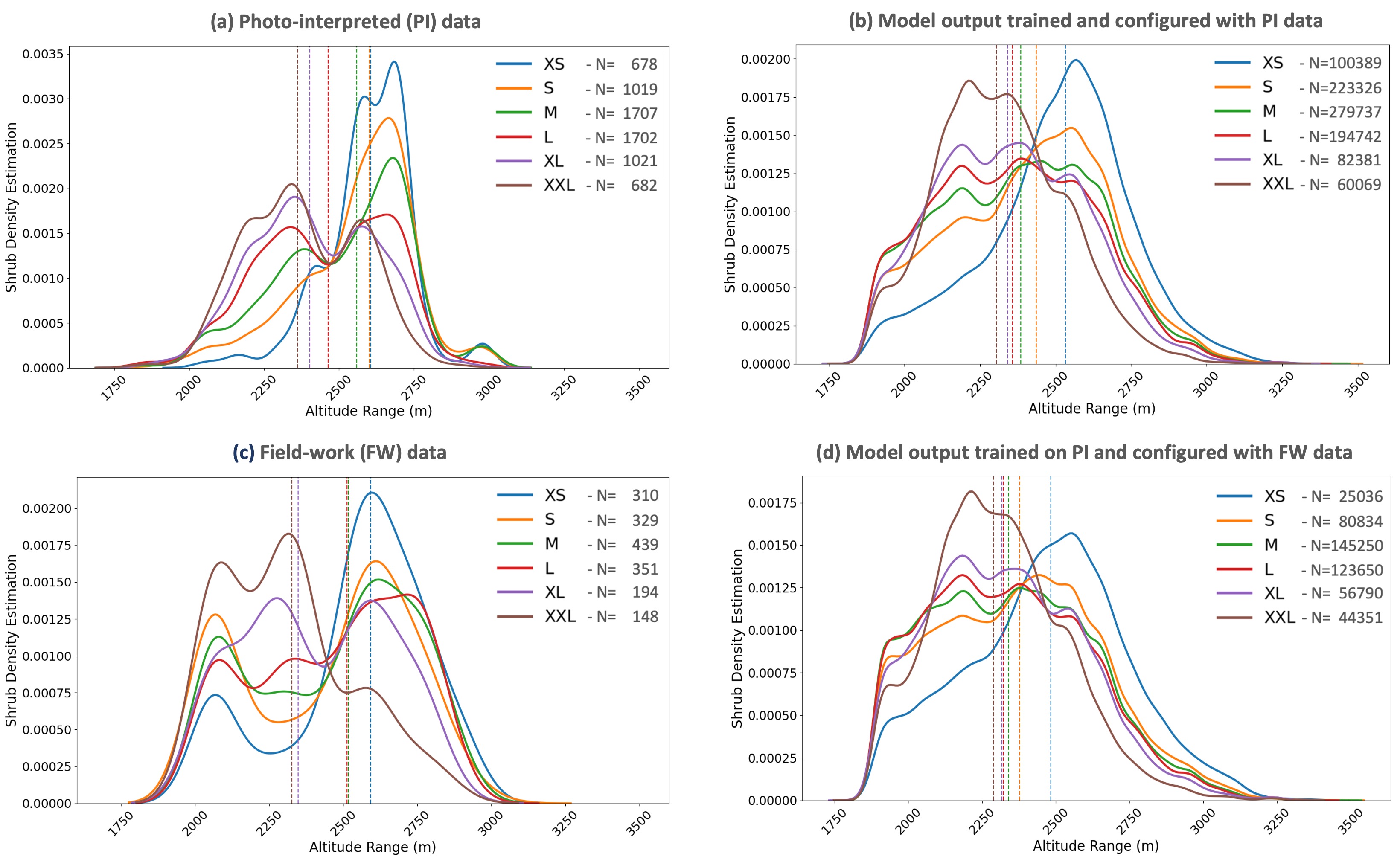} 
  \end{center}
  \caption{The distribution of junipers by size across different altitudes, comparing results from: (a) PI data samples, (b) model’s deployment output trained and configured with PI data, (c) FW data samples, and (d) model’s deployment output trained on PI data but configured with FW data. The model configuration refers to the selection of the confidence score threshold during deployment. N refers to the number of detected individuals and the dashed line to the median value.}
  \label{dist}
\end{figure}

Fig. \ref{dist} shows the distribution of juniper sizes across various altitudes under different conditions: (a) using PI data only, (b) the model’s deployment output trained and configured with PI data, (c) FW data only, and (d) the model’s deployment output trained on PI data but configured with FW data. The model configuration refers to the selection of the confidence score threshold during deployment.
According to the model outputs, junipers are concentrated between 2 km and 2.6 km in altitude following an altitudinal order: smaller shrubs (XS, S, and M) are more likely to grow at higher altitudes, while larger shrubs (XXL, XL, and L) tend to grow at lower altitudes.
This pattern in juniper distribution observed in the model outputs was also present in the PI and FW data, suggesting an altitudinal uplift in the optimal performance of the species.

\begin{figure}[H]
\centering
    \begin{center}
        \includegraphics[width=12cm]{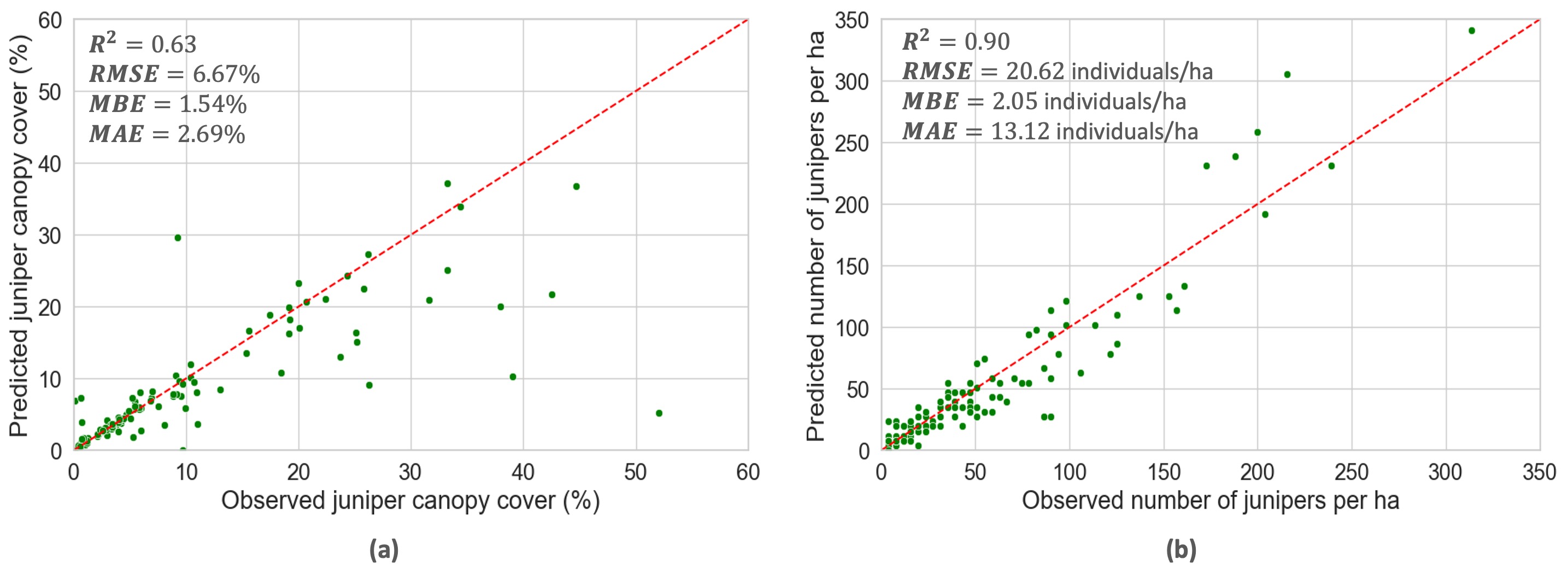}
    \end{center}

    \caption{Scatter plot of: (a) the percentage of area occupied by observed versus predicted junipers in each FW image. (b) The number of observed versus predicted junipers per hectare based on FW data.}
    \label{scatter}
\end{figure}

Fig. \ref{scatter}-a shows the scatter plot comparing the percentage of the FW image occupied by observed versus predicted junipers at a model confidence score threshold of $\theta_{score}=50\%$. There is a strong correlation between the two variables when junipers occupy a small portion of the image, with a Pearson correlation coefficient of $r=0.81$. However, this correlation weakens as the percentage of the image occupied by junipers increases. This suggests that the model segments junipers more accurately when they occupy a smaller portion of the image, and less accurately when they occupy a larger portion.
Fig. \ref{scatter}-b presents the scatter plot comparing the number of observed versus predicted junipers per hectare based on FW data at the same confidence threshold. There is a high correlation ($r=0.95$) when the image contains a small number of shrubs, but this correlation decreases as the number of shrubs in the image increases. This indicates that the model's estimation of juniper count is highly accurate when juniper density is low, but this accuracy diminishes as density increases.
Figure \ref{scatter} also assess the uncertainty in the generated juniper map in terms of canopy cover and shrub density: we found an error of $RMSE=20.62$, $MAE=13.12$, and $MBE=2.05$ individuals per hectare, and an error of $RMSE=6.67\%$, $MAE=2.69\%$, and $MBE=1.54\%$ for canopy cover.

\section{Discussion} \label{S5}



\subsection{Ecological considerations}

Our methodology, which combines AI and RS technologies with the new data construction design, was powerful in accurately delineating medium to large junipers from RGB satellite images within a complex mountain environment. Such a tool could be extended to systematically produce high-precision maps of juniper (or similar shrubs) in high-mountains or high latitudes to track climate change effects on their distribution, abundance, size structure, and in the woody-line throughout the Palearctic.

This new tool makes it possible to count all the individuals present in a large geographic region (e.g. a mountain range), determining their size with a precision of 13x13 $cm^2$.  It allowed us to quantify the distribution, abundance, and size (i.e., demographic structure) of individuals, which helped us to determine whether these population parameters varied as a function of environmental variables, such as altitude, in mountain environments. In short, it allowed us to do things that are not possible with traditional field methodologies.

Our analysis showed that junipers are more abundant in the North-West region within a specific range of altitude (mainly between 2km and 2.6km). Our results also revealed a massive difference in shrub size with altitude, i.e., the distribution of small shrubs is biased towards the highest altitudes while the largest ones tend to occur at the lowest altitudes. Such dominance of the smallest individuals in the highest altitudes could be indicative of a process of altitudinal rise as a consequence of global warming, which requires further investigation using historical aerial photography and field monitoring. If this were the case, during the last decades, juniper communis individuals would have been establishing more successfully at higher altitudes, where they found the preferred temperature range, and suitable habitats with less competition for resources with other plant species. Furthermore, the abandonment of traditional land uses (e.g. burning of juniper shrublands to increase the area of pastures for livestock \citep{lorite2001vegetacion, zamora2022managing}) has favored the recent recovery of the juniper cover. The obtained results can be used to investigate the factors explaining juniper distribution, and can further be employed by policymakers to establish efficient management plans for conservation and restoration in mountain areas.

\subsection{Methodological considerations}


We introduced a novel publicly available dataset of very-high resolution RGB satellite images. This is the first DL-ready dataset of juniper species, and the largest of its kind in which 8580 shrubs were digitized \citep{james2020detecting, retallack2022using}. This is the first dataset for individual delineation of polymorphic shrubs since all existing studies created datasets of plants with well defined and stable patterns, and a straightforward spatial distribution where most of them are private or have restricted access \citep{zheng2022deep, gan2023tree}.
Our primary objective is to present a valuable resource for designing and developing DL models to individually delineate medium to large shrubs in particular, and polymorphic plants in general from freely available high-resolution satellite imagery.

DL models are known as data-hungry models \citep{adadi2021survey}. However, collecting data through field surveying is costly, laborious, time-consuming, unsustainable, and spatially restricted. 
To the best of our knowledge, a data design handling such issues is lacking in the literature, as existing studies developed and validated their models only on FW data \citep{guirado2021mask, retallack2022using}. 
Thus, we proposed a new data construction approach (i.e., dual PI-FW data design) that consist in developing the model with PI data, then validating it using FW data. Our outcomes proved the efficiency and scalability of this approach in developing a delineation model in a more optimized way. 

We developed a new evaluation metric called S-IoU to assess the delineation performance of DL models, particularly for polymorphic species affected by overlapping, splitting due to canopy thinning, and uncertain human expert annotations. While all existing studies evaluating plant delineation models have relied on the IoU metric \citep{zheng2023surveying, gan2023tree}, our results demonstrate that S-IoU offers valuable insights into the proportion of a plant's area detected by the model and can be effectively used alongside IoU for a more comprehensive evaluation of DL models.

To the best of our knowledge, this is the first study exploring the potential of AI models to delineate individual plants with polymorphic nature since all existing studies handle the delineation of plants with well defined patterns \citep{zheng2023surveying, gan2023tree}.
It is also the first attempt to deploy a DL model at the regional scale for individual delineation of high-mountain shrubs \citep{guirado2021mask, retallack2022using}, simultaneously offering valuable insights into the distribution of these shrubs.

The proposed methodology offers valuable insights for the scientific community interested in using RS and AI technologies to delineate polymorphic plants individually, especially in situations where limited FW data is available. The generated data can be used to pretrain models for detecting similar types of shrubs in other high-mountain ecosystems or for identifying different shrub species.


\subsection{Limitations}
Our model provides useful insights about shrubs distribution in high mountains, however, it still makes some FPs and FNs. FPs can be manifested in the detection of other kinds of shrubs, isolated trees, small dark rocks with shades, edges of lagoons, and parts of "borreguiles" (i.e., humid pastures/grasslands). FNs can be noticed when shrubs grow in colonies, when their size is very large and covers a great proportion of the image tile, when they have low foliage densities, when the image background is very dark, and when the land cover is very patchy.

The reasons of these FPs and FNs can be summarized as follows:  
\begin{enumerate}

\item The temporal distribution of satellite images: this is related to the season in which the image was captured. Winter-based images are more likely to contain snow, and the soils are darker due to humidity, less sun light, and the existence of clouds and cloud shadows. 

\item The spatial distribution of satellite images: images are captured from different angles that can introduce  shadows from trees, hills, etc. Sometimes, they can have different resolutions where some areas are captured with very high resolution while others with less resolution. 

\item The spectral resolution of satellite images: in the case of this study, only three visual bands (RGB) are used. Adding more spectral information may provide enough information to help the model avoid confusing rocks, lagoons, and humid grasslads with shrubs. 

\item The local heterogeneity and patchiness of the area: areas with homogeneous land covers, such as barren lands, are easier for the model to detect shrubs than areas with high heterogeneity such as urban areas or cropland mosaics. 

\item The complex patterns of juniper: although we tried to make our data representative of the different shapes and backgrounds, it is difficult to collect all aspect in nature of this multiple faces shrub. 

\item The distribution of the PI dataset: given that the PI data relies on expert visual inspection of images, we excluded images featuring complex behavior due to low expert confidence. 

\end{enumerate}

To solve some of the aforementioned issues and further improve the model performance, ancillary information can be included such as orientation, altitude, and slope. For instance, stripe sized shrubs can only grow in areas with high slopes. Multispectral data can also be a good alternative to prevent the model from making false detections. In addition, other DL models architectures are encouraged for evaluation.

\section{Conclusion} \label{S6}
In this study, we digitized 8580 shrubs and demonstrated the potential of combining remotely sensed RGB imagery with Mask R-CNN model using a new data construction design to individually segment medium to large juniper shrubs in high-mountain ecosystems from freely available high-resolution satellite imagery.

Our deployment results revealed that these shrubs exhibit a pronounced concentration in the North-West region, within a specific range of altitude where smaller shrubs tend to occur at higher altitudes, while larger shrubs are more prevalent at lower altitudes. Such potential shift in the altitudinal range will be investigated in further research. Our work and cartography will assist to the management and conservation of the Sierra Nevada National Park, a global hotspot for biodiversity, in the face of global warming.

Our experimental results highlight the effectiveness of the proposed dual data construction approach in addressing the limitations of traditional field surveying methods. They also demonstrate the robustness of the developed S-IoU metric, making it a valuable complement to IoU for evaluating DL models on polymorphic plants. Additionally, the results showcase the potential of Mask R-CNN in segmenting polymorphic plants, achieving F1-scores of 87.87\% and 76.86\% in the PI and FW test sets, respectively

While this study marks a significant advancement in the application of RS and DL for individual shrub delineation, it is important to acknowledge its limitations. The delineation accuracy is influenced by several factors, including the spatial-temporal distribution of satellite images, the spectral resolution of satellite images, the heterogeneity of the background, the polymorphic nature of junipers, and the distribution of the PI data. Future studies should focus on refining these aspects to achieve even greater accuracy by using ancillary information (topographic, atmospheric, etc.), multispectral data, and new delineation models.

\section*{Declaration of Competing Interest}
The authors declare that they have no known competing financial interests or personal relationships that could have appeared to influence the work reported in this paper.

\section*{Data availability}
This dataset (Version 0.0.1) \citep{khaldi_2024_10058715} is available to the public through an unrestricted data repository hosted by \href{https://zenodo.org/}{Zenodo} at: \blue{\url{https://zenodo.org/records/10058715}}

\section*{Acknowledgements}
This work was part of the project SmartFoRest (TED2021-129690B-I00, funded by MCIN/AEI/10.13039/ 501100011033 and by the European Union NextGenerationEU/ PRTR). It was initially supported by project DETECTOR (A-RNM-256-UGR18 Universidad de Granada/FEDER) and LifeWatch-ERIC SmartEcomountains (LifeWatch-2019-10-UGR-01 Ministerio de Ciencia e Innovación/Universidad de Granada/FEDER), and it was part of DeepL-ISCO (A-TIC-458-UGR18 Ministerio de Ciencia e Innovación/FEDER), and BigDDL-CET (P18-FR-4961 Ministerio de Ciencia e Innovación/Universidad de Granada/FEDER).

We sincerely acknowledge Javier Cabello and Cecilio Oyonarte from the Andalusian Center for the Assessment and Monitoring of Global Change of the University of Almeria for their help during the field work and for sharing their life-long experience on Sierra Nevada ecosystems, which contributed to frame this work. We also thank Irati Nieto and Carlos Navarro for their help during field works.

\bibliography{sample}

\appendix
\section[\appendixname~\thesection]{Shrub size categorization}

Notably, juniper exhibits a significant variation in its size (Table \ref{tab2}). 
As a result, we categorized the shrubs into six distinct groups based on their size (Table \ref{tab0}): extra small (XS), small (S), medium (M), large (L), extra large (XL), and extra extra large (XXL). This was done using the PI data quantiles, since it is well distributed than the FW data. 

\begin{table}[H]
\centering
\caption{Statistical description of juniper per square meters.}
\label{tab2}
\begin{tabular}{lcrrrrrr}
\hline

\multirow{2}{*}{\textbf{\thead{Data name}}} & \multicolumn{6}{c}{\textbf{\thead{Statistics per size ($m^2$)}}} \\
 & \textbf{\thead{Mean}} & \multicolumn{1}{c}{\textbf{\thead{Std}}} & \multicolumn{1}{c}{\textbf{\thead{Min}}} & \multicolumn{1}{c}{\textbf{\thead{25\%}}} & \multicolumn{1}{c}{\textbf{\thead{75\%}}} & \multicolumn{1}{c}{\textbf{\thead{Max}}} & \multicolumn{1}{c}{\textbf{\thead{Count}}}\\
\hline
Photo interpreted & \multicolumn{1}{r}{19.97} & 42.39 & 0.13 & 3.62 & 20.82 & 761.42 & 6809 \\ 
Field work & \multicolumn{1}{r}{15.99} & 38.75 & 0.16 & 2.36 & 16.42 & 970.60 & 1771\\
\hline
\end{tabular}%
\end{table}

\begin{table}[H]
\centering
\caption{Description of shrub size classification schema.}
\label{tab0}
\begin{tabular}{ccccccc}
\hline
 & \multicolumn{6}{c}{Shrub size categories} \\ 
 \hline
 & XS & S & M & L & XL & XXL \\
 \hline
Range ($m^2$) & $[0.13, 1.72[$ & $[1.72, 3.62[$ & $[3.62, 9.08[$ & $[9.08, 20.82[$ & $[20.82, 41.06[$ & $[41.06, \infty[$ \\

Quantiles (\%) & $0-10$ & $10-25$ & $25-50$ & $50-75$ & $75-90$ & $90-100$ \\

\hline
\end{tabular}
\end{table}


\section[\appendixname~\thesection]{Soft version of IoU metric} \label{B}
\begin{figure}[H]
  \begin{center}
    \includegraphics[width=10cm]{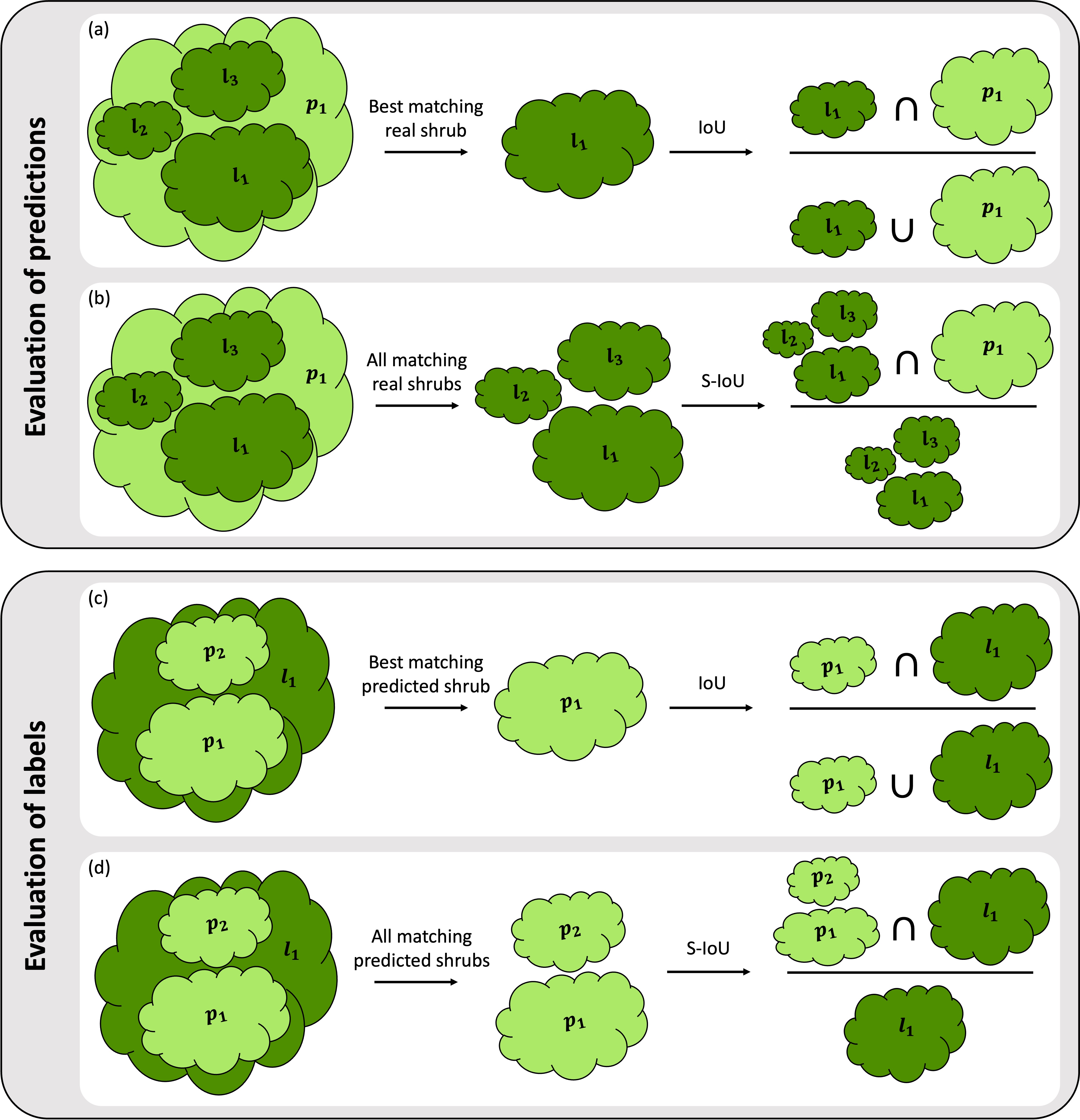} 
  \end{center}
  \caption{Description of the difference between IoU and MIoGTA metrics for predictions and labels evaluation. 
  $p$ denotes the model predictions (light green) and $l$ the expert annotations (dark green).}
  \label{IOU_vs_MIOGTA}
\end{figure}

We develop a soft version of IoU that we name S-IoU. 
The main concept behind this metric is to count all matching shrubs in the evaluation of labels and predictions (Fig \ref{IOU_vs_MIOGTA} (b),(d)), rather than focusing solely on the best-matching shrub as performed by IoU metric (Fig \ref{IOU_vs_MIOGTA} (a),(c)).

\end{document}